\documentclass[english,twocolumn,10pt,letterpaper]{article}

\usepackage{helvet} % DO NOT CHANGE THIS
\usepackage{courier}  % DO NOT CHANGE THIS
\usepackage[hyphens]{url}  % DO NOT CHANGE THIS
\usepackage{graphicx} % DO NOT CHANGE THIS
\usepackage{graphicx}  % DO NOT CHANGE THIS
\makeatletter

\usepackage{courier}
\usepackage[T1]{fontenc}
\usepackage[utf8]{inputenc}
\usepackage{verbatim}
\usepackage{float}
\usepackage{mathtools}
\usepackage{amsmath}				
\usepackage{amssymb}
\usepackage{breakurl}
\usepackage{rotating}
\usepackage{multirow}
\usepackage{booktabs}

\usepackage{cvpr}

\usepackage{times}
\usepackage{array}
\usepackage{verbatim}
\usepackage{multirow}
\usepackage{rotating}
\usepackage{amsthm}
\usepackage{xcolor}
\usepackage[nointegrals]{wasysym}
\usepackage{url}
\usepackage{algorithm}
\usepackage{algpseudocode}
\usepackage{amsfonts}
\usepackage{xspace}

\usepackage[breaklinks=true,bookmarks=false]{hyperref}

\cvprfinalcopy % *** Uncomment this line for the final submission

\def\cvprPaperID{6262} % *** Enter the CVPR Paper ID here

\ifcvprfinal\fi
\newcommand{\action}{q}
\newcommand{\answer}{a}
\newcommand{\actiont}{\action^{(t)}}
\newcommand{\answert}{\answer^{(t)}}

\newcommand{\answertopt}{\answert} % {\answer_*^{(t)}}
\newcommand{\state}{\mathbf{s}}
\newcommand{\states}{\mathcal{S}}
\newcommand{\statet}{\state^{(t)}}
\newcommand{\statetpo}{\state^{(t+1)}}

\newcommand{\rwt}{r(\statet,\actiont)}
\newcommand{\rwtnew}{r^{\text{new}}(\statet,\actiont)}
\newcommand{\btheta}{{\boldsymbol\theta}}
\newcommand{\bomega}{{\boldsymbol\omega}}
\newcommand{\bphi}{\state'} %\boldsymbol{\phi}}
\newcommand{\bvarphi}{\state'} %\boldsymbol{\varphi}}
\newcommand{\bPhi}{\states} %\boldsymbol{\Phi}}
\newcommand{\pans}{p} %^{\uparrow}}
\newcommand{\pnoans}{p} %^{\downarrow}}
\newcommand{\ppi}{{\pi}} % p_pi
\newcommand{\ppip}{{\pi_{0}}} % p_pi prior

\newcommand{\kl}{\mathbf{KL}}

\newcommand{\actions}{\mathbf{\mathcal{Q}}}
\newcommand{\answers}{\mathbf{\mathcal{A}}}

\newcommand{\hist}{C}
\newcommand{\histT}{C^+}

\newcommand{\Y}{O}
\newcommand{\y}{o}
\newcommand{\ystar}{o^*}

\newcommand{\E}{\mathbb{E}}
\newcommand{\Eemp}{\hat{\E}}

\newcommand{\gain}{\mathcal{G}}

\newcommand{\ansr}{\textsc{responder}\xspace}  %answerer
\newcommand{\Ansr}{\textsc{Responder}\xspace}
\newcommand{\enc}{\textsc{encoder}\xspace}  %answerer

\newcommand{\seeker}{\textsc{seeker}\xspace}
\newcommand{\Seeker}{\textsc{Seeker}\xspace}
\newcommand{\exec}{\textsc{executor}\xspace}

\@ifundefined{showcaptionsetup}{}{%
 \PassOptionsToPackage{caption=false}{subfig}}
\usepackage{subfig}
\makeatother

\usepackage{babel}

\title{Gold Seeker: Information Gain from Policy Distributions \\
for Goal-oriented Vision-and-Langauge Reasoning}

\begin{document}

 \author{Ehsan Abbasnejad$^1$, Iman Abbasnejad$^2$, Qi Wu$^1$, Javen Shi$^1$, Anton van den Hengel$^1$ \\
 $^1$\texttt{\small{}\{ehsan.abbasnejad,qi.wu01,javen.shi,anton.vandenhengel\}@adelaide.edu.au} \\
 $^2$\texttt{\small{}{i.abbasnejad@fugro.com}} \\
  $^1$Australian Institute for Machine Learning \&
  The University of Adelaide, Australia, \\$^2$Fugro Australia Marine
 }

\maketitle

\begin{abstract}

As Computer Vision moves from passive analysis of pixels to active analysis of semantics, the breadth of information algorithms need to reason over has expanded significantly.
%This is particularly visible in vision-and-language problems as they may need to reason over, and interact with a human regarding, widely varying image content, questions and answers.
One of the key challenges in this vein is the ability to identify the information required to make a decision, and select an action that will recover it.  We propose a reinforcement-learning approach that maintains a distribution over its internal information, thus explicitly representing the ambiguity in what it knows, and needs to know, towards achieving its goal.  Potential actions are then generated according to this distribution.  For each potential action a distribution of the expected outcomes is calculated, and the value of the potential information gain assessed.  The action taken is that which maximizes the potential information gain.
%Aobtained, as compared to the existing internal information. 
We demonstrate this approach applied to two vision-and-language problems that have attracted significant recent interest, visual dialog and visual query generation. In both cases the method actively selects actions that will best reduce its internal uncertainty, and outperforms its competitors in achieving the goal of the challenge.

%OR

%Identifying the gaps in one's knowledge is complex at best.  Identifying the actions required to fill these gaps is even more challenging.  We propose an approach applicable to reinforcement learning agents that maintains an explicit distribution over the internal information held, and uses it to identify the action the agent should take to acquire the knowledge it needs to achieve its goal.  The method hypothesises multiple actions, and predicts a distribution of answers for each.  This allows it to select the action that is most likely to lead to the information required to achieve its goal, given its current state.  We demonstrate the application of this approach to two problems that require active reasoning over complex and varied information, visual dialogue and visual query generation.  The method actively selects actions that provide the information it needs to achieve its final goal, and outperforms its comparators in doing so.

\end{abstract}

\section{Introduction}

% In most traditional problems in computer vision it is assumed that all of the information required is available a-priori and suitable to be embodied in the code or the weights of the solution.  This assumption is so pervasive that it typically goes unsaid. In fact, this assumption is satisfied by a small subset of problems of practical interest.  Problems in this set must be self-contained, tightly specified, relate to a very prescribed form of data drawn from a static distribution, and be completely predictable. Many important problems do not meet these criteria, though researchers have found many that do.

The majority of problems that computer vision might be applied to greatly benefit from agents capable of actively seeking the information they need.  This might be because the information required is not available at training time, or because it is too broad to be embodied in the code or weights of an algorithm.
The ability to seek the information required to complete a task enables a degree of flexibility and robustness that cannot be achieved through other means.

% \begin{figure}
%     \centering\includegraphics[width=0.9\columnwidth]{img/fig1_new1.pdf}
%     \caption{
%     %A Goal-oriented vision-and-language task, which consists of four parts:
%     Two goal-oriented vision-and-language tasks, broken up into four constituent parts: a context encoder, an information seeker, a responder and a goal executor. The given examples are from a visual dialog dataset GuessWhat \cite{guesswhat_game} (above the red dash line) and, a compositional VQA dataset CLEVR \cite{clevr} (below).
%     }
%     \vspace{-10pt}
%     \label{fig:intro}
% \end{figure}
\begin{figure}
\vspace{0mm}
    \centering\includegraphics[width=0.9\columnwidth]{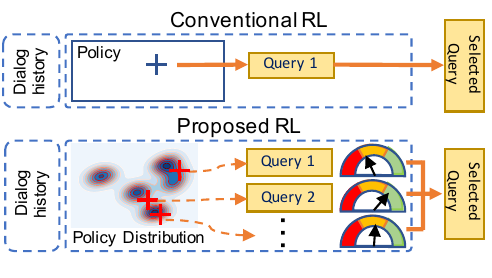}\vspace{-4mm}
    \caption{
    %A Conventional information Seekers map the dialog history along with the input image to a policy \emph{point} from which the query is generated. Our Seeker however, learns a \emph{distribution} over policies. Each sample from this distribution produces a query that a \emph{responder} evaluates the gain from knowing its answer.
		\small{The role of the \seeker in many conventional RL-based agents is to select an action based on the current (single) policy.  In goal-oriented visual dialogue this means selecting the next query based on the image and the dialogue thus far.  Our proposed \seeker instead exploits a distribution of policies to generate multiple query hypotheses. The one that maximises the potential information gain is then selected.  In this sense the agent is able to identify gaps in the information it holds and ask questions that will fill them.}
    }\vspace{-2pt}
    \label{fig:intro}
\end{figure}

Some of the applications that lie at the intersection of vision and language have this property, including visual dialog \cite{das2016visual,visdial_rl}, visual question answering \cite{guesswhat_game,answer_questioner_mind,strub2017end}, and vision-and-language navigation\cite{anderson2018vision}.
These problems require an agent (model) to acquire information on the fly to help to make decisions, because the space of all possible questions (or dialogues) encompasses more information than can be encoded in a training set.
Additionally, a range of tasks have been proposed recently that use `language generation' as a mechanism to gather information
towards achieving a (non-language based) goal\cite{tempered,answer_questioner_mind,lee2018answerer}. These tasks offer a particular challenge because the set of all information that might possibly be involved is inevitably very broad, which makes concrete representations difficult to employ practically.

In a visual dialog, and particularly goal-oriented visual question generation,
an agent needs to understand the user
request and complete a task by asking a limited number of questions.
Similarly, compositional VQA (e.g. \cite{clevr}) is a visual query generation problem that requires
a model first to convert a natural language question to a sequence
of actions (a `program') and then obtain the answer by running the programs on
an engine. The question-to-program model represents an information `seeker', while the broader goal is to generate an answer based on the information acquired.

\begin{figure}
	\centering\includegraphics[width=1\columnwidth]{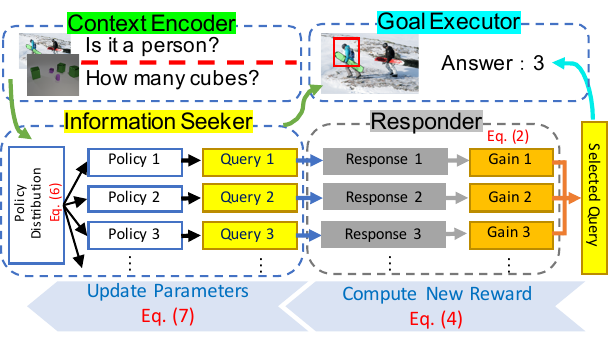}\vspace{-5pt}
	\caption{
		Information seeker maintains a distribution over the policies. Each sample (particle) from the distribution can generate a different query. For each query, responder evaluates its \emph{gain} and the best one is chosen to be executed in \exec. Based on how well the query leads to achieving its goal, the \emph{policy distribution} is updated.
	}\label{fig:framework}\vspace{-2pt}
\end{figure}

Agents applicable to these tasks typically consist of four parts: a \textit{context encoder}, an \textit{information seeker}, a \textit{responder} and a \textit{goal executor}, as shown in Fig.\ref{fig:framework}. The context encoder is responsible for encoding information such as images, questions, or dialog history to a feature vector. The information seeker is a model that is able to generate new queries (such as natural language questions and programs) based on the goal of the given task
and its strategy. The information returned by the responder is added to the context and internal information and
sent to the goal executor model to achieve the goal.
The seeker model plays a crucial role in goal-oriented vision-and-language tasks, as
better seeking strategies that recover more information
improve the chance of the goal being achieved.
Moreover, the seeker's knowledge of the value of additional information
is essential in directing the seeker towards querying what is needed to
achieve the goal.
In this paper, we focus on exploring the \textbf{seeker} and \textbf{responder} models.

The conventional `seeker' models in these tasks follow a
sequence-to-sequence generation architecture, that is, they translate an image
to a question, or translate a question to a program sequence via supervised
learning.  This requires large numbers of ground-truth training pairs.
Reinforcement learning (RL) is thus employed in such goal-oriented vision-language
tasks to mediate this problem due to the RL's ability to focus on achieving a goal through directed trial and error~\cite{guesswhat_game}. A policy in RL models specifies how
the seeker asks for additional information.
However, these methods generally suffer from
two major drawbacks:
(1)  they maintain a single policy that
translates the input sequence to the output while disregarding the
strategic diversity needed. Intuitively a single policy is not enough
in querying diverse information content for various goals--we need multiple strategies.
In addition, (2) the RL employed in these
approaches can be prohibitively inefficient since the question generation process (or query generation)
does not consider its effect in directing the agent towards the goal.
In fact, the agent does not have a notion of what information it needs
and how it benefits in achieving its goal.

To this end, in contrast to conventional methods that use a single policy to model a vision-and-language
task, we instead maintain a \textbf{\emph{distribution of policies}}.
By employing a Bayesian reinforcement learning framework for learning this
distribution of the seeker's policy, our model incorporates
the expected gain from a query towards achieving its goal.
Our framework, summarized in Fig. \ref{fig:intro}, uses recently proposed Stein Variational
Gradient Descent \cite{svgd} to perform an efficient update of
the posterior policies.  Having a distribution over seeking policies, our agent is \textbf{\textit{capable
of considering various strategies for obtaining further information}},
analogous to human contemplation of various ways to ask a question.
Each sample from the seeker's policy posterior represents a policy of its own,
and seeks a different piece of information.
This allows the agent to further contemplate the outcome of the various
strategies before seeking additional information and considers
the consequence towards the goal. We then formalize an approach for the agent
to \textbf{\textit{evaluate the consequence of receiving additional information}} towards
achieving its goal. This consequence is the intrinsic reward the policy distribution receives to update its strategy for question generation.

We apply the proposed approach to two complex vision-and-language
tasks, namely GuessWhat \cite{guesswhat_game} and CLEVR~\cite{clevr},
and show that it outperforms the comparative baselines and achieves the state-of-art results.

\section{Related work}
\paragraph{Goal-oriented Visual Dialog}
%AVisual dialog is a recently proposed vision-and-language task that began with image captioning \cite{Karpathy2014deepvs,vinyals2014show,wu2015image} and, includes visual question answering \cite{antol2015vqa,ren2015image,wu2015ask}. 
%/A
Das \etal \cite{das2016visual} proposed a visual dialogue task that requires an agent to engage in conversation with a human, centred on the content of a given image. They further (in~\cite{visdial_rl}) propose the use of reinforcement learning in two tasks for visual dialog.
%Das \etal \cite{das2017learning} establish two reinforcement learning based agents corresponding to question and answer generation respectively, to finally locate an unseen image from a set of images. The question agent predicts the feature representation of the image and the reward function is given by measuring how close the representation is compared to the true feature.  
de Vries \etal in
\cite{guesswhat_game} propose a Guess-What game dataset,
where one person asks questions about an image to guess which object
has been selected, and the second person answers. 
%A
This was a critical step in the development of goal-oriented visual dialogue because the objective there was not merely to continue the conversation, but rather to achieve a secondary goal (winning the game).
%/A
Lee \etal \cite{lee2018answerer} then developed an information theoretic approach which allows a questioner to ask appropriate consecutive questions in GuessWhat.

% \vspace{-10pt}
% \paragraph{Compositional VQA}
% An increasingly popular research direction in visual question answering is to consider modular architectures. This approach involves connecting distinct modules designed for specific desired capabilities such as memory, or specific types of reasoning. Neural Module Networks (NMNs) were introduced by Andreas \etal in \cite{andreas2016learning,andreas2015deep}.
% There the question parse tree is turned into an assembly of modules from a predefined set, which are then used to answer the question. Johnson \etal \cite{clevr} propose a Compositional Language and Elementary Visual Reasoning (CLEVR) dataset that allows the question to be transferred to a sequence of functional problems, which can be further used to query information for the structured scene representation to help to answer the question. In this paper, our  `seeker' model is used as a program generator to generate functional programs from questions.

\vspace{-10pt}
\paragraph{RL in Vision-and-Language problems}
Reinforcement learning (RL) \cite{kaelbling1996reinforcement,sutton1998reinforcement} has been adopted in several vision-and-language problems, including image captioning~\cite{liu2016optimization,ren2017deep,rennie2016self}, VQA~ \cite{hu2017learning}, and visual dialogue~\cite{visdial_rl,lu2017best}. Recently, some works \cite{asghar2016online,su2016continuously}
have integrated Seq2Seq models and RL. RL has also
been widely used to improve dialogue managers, which manage transitions
between dialogue states \cite{pietquin2011sample}. 
%Most recently, Wu~\etal \cite{wu2017you} combines reinforcement learning and generative adversarial networks (GANs) to generate more human-like dialogues. 
However, nearly all of the methods use a single policy that translates the input sequence to an output. % while disregarding the strategic diversity needed.
In our work, we instead maintain a distribution of policies.

\vspace{-10pt}
\paragraph{Intrinsic rewards}
Intrinsic rewards refer to rewards beyond those gained from the environment in RL. These rewards are motivated by the sparse nature of environmental rewards and a need to motivate better exploration.
For example, curiosity \cite{curiosity} is one such intrinsic reward mechanism by which agents are encouraged to visit new states.  This idea has been extended to employ Bayesian methods to learn the expected improvement of the policy for taking an action \cite{vime,bayesian_surprise}.
We use the expected gain in a vision-and-language task as an intrinsic reward to improve our model.
% Nevertheless, our approach is flexible enough that can be easily integrated with any of such additional intrinsic rewards.

\section{Goal-oriented Vision-Language Task}

We represent our goal-oriented vision-and-language solution as having four constituent parts. The \seeker takes as input the encoded image and context features produced by an \enc to generate a query to seek more information from a \ansr\!\!, that will generate a response. 
%A*** check the following 
The role of \ansr is to model the environment in order to allow the agent to determine which query is best to ask. It takes in a question to produce an answer and its predicted score. Our \ansr is an extension of the 'A-\textsc{bot}' in \cite{visdial_rl}, in that it instead considers a distribution over possible answers and generates a corresponding upper bound on the score of the question.  This is a critical distinction, and represents an approach that is far more statistically justifiable. To enable our approach we develop a synthetic \ansr inspired by the neuro-scientific formulation of agency that computes the translation of intentions into actions and evaluates their consequences in terms of predicted and actual experiences \cite{comparator}.
%This process can be performed multiple rounds until the \seeker gathered enough information.
%A The alternative queries and their potential responses according to the \ansr are evaluated and the query that carries most information is chosen by the \seeker\!\!.The queried information is sent to the fourth agent, the \exec that makes the final prediction. The game is recognized as a success if the prediction hits the given goal.

Typically the \seeker learns a policy to generate queries on the basis of image and context features.
The primary novelty in our proposed approach is that the \seeker instead maintains a distribution of policies that enables multiple query hypotheses to be sampled.  The \ansr then calculates an upper bound on the information gain for each hypothesis.  The final query is that corresponding to the maximal (upper bound on the) information gain.
%/A

%for a collection of items.

%The items are either
%1) multiple objects in an image for one agent to identify the unknown
%object of interest of the other agent by asking questions, or 2) objects
%for the agents to split by negotiation. Conditioned on this game,
%once enough information is gathered a \emph{Guesser} takes the dialogue
%history and predicts the goal. The game is a success when the goal
%is achieved. The game between these two agents effectively simulates
%the real natural language based conversation to achieve a particular
%goal, e.g. uncovering an unknown object, or an agreed split.

Formally, for each game at round $t$, we have a tuple $(I,\hist,\actiont)$, where $I$
is the observed image, $\hist$ is the context information at the current round\footnote{The dependency on $t$ is dropped for clarity.} and $\actiont$ is a query generated by the \seeker agent. Subsequently, $\actiont$ is sent to the \ansr that generates a response $\answert$. The \ansr imitates the potential response for a query and evaluates its value.
After $T$ rounds of this `seek-answer' process, the tuple $(I,C,\{\actiont\}_{t=1}^T,\{\answert\}_{t=1}^T)$ is sent to the \exec who selects the target from the candidate list $\Y=\{o_1,o_2,...,o_N\}$. The ground truth target is denoted as $\ystar$ and the game is success if the $\ystar$ is successfully selected by the \exec.

To be more specific, in the Guesswhat (visual dialog) setting, $\hist$ is the dialog history and $\actiont$ is a natural language question. Then, $\Y$ is the candidate object bounding boxes. The answer $\answert$ is provided by the oracle as either Yes/No or N/A (not applicable for the cases when the question is unrelated). 
In the CLEVR (VQA), $\hist$ is a single question asked by users and $\actiont$ is a functional program, while the $\Y$ is a candidate answer vocabulary. The answers in this problem are the same as the target candidates.
% In this paper, we adapt pre-trained, fixed \enc, \ansr and \exec in the game and only focus on training a better \seeker, which will be discussed in the following section.

% \section{Preliminaries}

% We here provide some background in reinforcement learning and discuss the
% policy gradient estimation that we use in the paper. %method that we will modify for a contemplation-based question generation. 

\subsection{Reinforcement Learning}

Reinforcement learning considers agents interacting with their environment
by taking a sequence of actions and assessing their effect through a scalar reward. 
The agent's task is to learn a \emph{policy} that maximizes the expected cumulative rewards from its interaction with the 
environment.

Consider a vision-and-language task where the agent generates a query $\actiont\in\actions$
at each time step $t$ given the state $\statet$. Each $\statet$ encompasses
the history of the dialog (including past query-answer pairs) and
the input image. Upon receiving an answer $\answert\in\answers$ for the query,
the agent then observes a new state  
$\statetpo$ 
and receives a
scalar reward $\rwt\in R$. The goal of the reinforcement 
learning in this task is to find a querying policy $\ppi(\actiont|\statet,\btheta)$
given the state $\statet$ to maximize an expected return:
{\small\begin{equation}
J(\ppi)=\E_{\state_{0},\action_{0},...\sim\ppi}[\sum_{t=0}^{\infty}\gamma^{t}r(\statet,\actiont)], \nonumber
\end{equation}}
where 
$0\le\gamma^t\le1$ 
is a discount factor. State variable $\statet$ is generally considerred to encompass all the 
information needed for the agent to take an action (in our application, generate a query). 
The expected return $J$ depends on $\ppi$ because $\actiont\sim\ppi(\actiont|\statet, \btheta)$
drawn from the policy (distribution) $\ppi$ (\ie $\ppi(\btheta|\hist,I,C)$% where $D_t=\{\{\action^{(t'-1)}\}_{t'=1}^{t-1},\{\answer^{(t')}\}_{t'=1}^{t-1}\}$)
. The state $\state^{(t+1)}\sim P(\state^{(t+1)}|\statet,\actiont)$
is generated by the seeker's environmental dynamics which are unknown.
%The state value function 
%$
%V^{\ppi}(\statet)=\E_{\answert,\state^{(t+1)},...\sim\ppi}[\sum_{i=0}^{\infty}\gamma^{i}r(\state^{(t+i)},\action^{(t+i)})]
%$
%is the expected return by policy $\ppi$ from state $\statet$. 
In policy gradient algorithms \cite{policy_grad} such as the well-known
REINFORCE \cite{reinforce}, the gradient 
is estimated by samples from the policy $\ppi(\action|\state,\btheta)$. Specifically, REINFORCE uses the
following approximator of the policy gradient: 
$$\nabla_{\btheta}J(\btheta)\approx\sum_{t=0}^{\infty}\nabla_{\btheta}\log\ppi(\actiont|\statet,\btheta)\rwt,$$
% {\small\begin{equation}\vspace{-3mm}
% \nabla_{\btheta}J(\btheta)\approx\sum_{t=0}^{\infty}\nabla_{\btheta}\log\ppi(\actiont|\statet;\btheta)\rwt\label{PG}
% \end{equation}}
This gradient is computed based on a single rollout trajectory, where $\rwt=\sum_{i=0}^{\infty}\gamma^{i}r(\state^{(t+i)},\action^{(t+i)})$
is the accumulated return from time step $t$.

\section{Information \Seeker and the \Ansr}
\label{sec:infoseeker_oracle}
As discussed, our approach maintains a distribution of policies, $\ppi$ and updates it using the \ansr and \exec.
% Our framework thus has three significant aspects: 
% (i) the \ansr  models belief over the potential answers and calculates an upper bound on the information gain for each potential query; % (Sec. \ref{subsec:answer});
% (ii) the \seeker models belief over the policy space rather than maintaining only a single policy. % (Sec. \ref{subsec:question});
% %A In sharp contrast to existing methods, we are interested in not only finding the right policy, but rather we model a multi-modal distribution of policies, to enable learning diversity seeking policies in analogy to human contemplation using multiple strategies.
% (iii) the \seeker updates its belief over the distribution of
% the policies by incorporating the feedback from the environment. % (Sec. \ref{subsec:refinement}).
In a nutshell, our approach takes the following steps for training an information seeking agent:
\begin{enumerate}
	\item Conditioned on the history and context, the distribution of the query is: 
	$$\ppi(\actiont|\statet,\hist,I) = \int\ppi(\actiont|\statet,\btheta)\ppi(\btheta|\hist,I)d\btheta,$$
	where we can sample to generate a query, \ie {\begin{eqnarray}\vspace{0mm}
	\actiont_i\sim \ppi(\actiont|\statet,\btheta_i),\quad \btheta_i\sim \ppi(\btheta|\hist,I)\label{eq:sample_act}\\
	\qquad\text{for}\quad i=1,\ldots,n; \nonumber
	\end{eqnarray}}where $n$ is the number of query samples simulating the alternative queries that could be made.
	% where $\ppi$ is the posterior of the policy distribution (see Section \ref{subsec:refinement})
	\item Our \ansr  models belief over the potential answers and calculates the gain for each query $\actiont_i$. Since ultimately we need to choose one query, we choose the one with highest gain and incorporate it into the reward for the RL (see Section \ref{subsec:answer});
	\item The \seeker models belief over the policy space rather than maintaining only a single policy (hence we can sample multiple parameters from its distribution in Eq.~(\ref{eq:sample_act})). The posterior $\ppi(\btheta|\{\actiont\}_t^T,\{\answertopt\}_t^T,o,\hist,I)$  considering the outcome of executing the query (potentially at multiple rounds) and a prior is formulated as part of the RL framework. Here, $\answertopt$ is the correct answer obtained from the environment. For example in case of GuessWhat game, it is the answer obtained from the oracle. (see Section \ref{subsec:question});
	\item The \seeker updates its belief over the distribution of
the policies by incorporating the feedback from the environment. This update has to ensure the posterior for the parameters of the \seeker remains valid (see Section \ref{subsec:refinement}).
\end{enumerate}

\subsection{\textbf{Query Gain and the \Ansr\label{subsec:answer}}}
In our approach the agent keeps a model (\ansr) of the environment to be able to predict what might most valuably be asked.
The agent uses this model to imitate
the behavior of the goal \exec and anticipate its potential response. Utilizing this model, the agent generates queries
with answers that bring it closer to achieving its goal. In particular, 
we define the \emph{gain} from state $\statet$ is,
 %. Thus we define the potential gain at state $\statet$ as:
% \begin{align}
% \gain_\bomega(\statet,\actiont) & = \E_{\answer}[u(\ansopt|\actiont,\statet,\hist,I,\bomega)]
% \nonumber \\
% &\qquad\qquad\qquad-\E_{\answer}[u(\answer|\actiont,\statet,\hist,I,\bomega))]\nonumber \\
% &=U_{\ansopt,\bomega}(\statet,\actiont)-U_{\bomega}(\statet,\actiont)\label{eq:ans_gain}
% \end{align}
\begin{align}
\gain_\bomega(\statet,\actiont) & = \E_{\answer}[u(p(\answer|\statet,\actiont,\hist,I;\bomega))]
% &\qquad\qquad\qquad \nonumber 
% &=U_{\ansopt,\bomega}(\statet,\actiont)-U_{\bomega}(\statet,\actiont)
\label{eq:ans_gain}
\end{align}
where $u$ is a scoring function for the answers and $\bomega$ is the set of parameters of the \textsc{responder}. 
Here, $p(\answer|\statet,\actiont,\hist,I;\bomega)$ is the probability of the answer for a given query in the \ansr. This gains effectively evaluates the score of an answer.
% In addition, $\bomega$ is the set of parameters of the \textsc{responder}.
Particularly we find $\bomega$ such that the expected goal under this policy
%$\E_{\answer\sim p(\answer|\statet,\hist,I;\bomega)}[p(\y|\answer,\statet;\bomega)]$ 
is maximized. Intuitively, the agent queries $\actiont$ at time $t$ only if it believes the answer $\answert$
it receives ultimately maximizes the {gain} in achieving its
goal at state $\statet$. 
For instance, in the GuessWhat game the \ansr takes in the history of the dialog and the current question and evaluates how good it is for achiveing the goal (\ie guessing the correct object).

In order to integrate this measure into an RL framework,
we use this gain in the reward. 
This reward is collected by choosing the best query according to its gain in Eq. (\ref{eq:ans_gain}), % eq:ucb_bound}),
i.e. $\max_{\actiont}{\gain}_\bomega(\statet,\actiont)$ (although in practice with probability $\epsilon$ we 
sample an alternative query from the \seeker's distribution $\ppi$ to encourage exploration).
Moreover, inspired
by curiosity-driven and information maximizing exploration \cite{vime,curiosity}, we
incorporate this gain as an intrinsic motivation to consider the gain, i.e.
{\begin{eqnarray}
\rwtnew\hspace{-2mm} & = &\hspace{-2mm}\rwt+\eta{\gain}_\bomega(\statet,\actiont)\label{eq:newreward} \\
J(\btheta)\hspace{-2mm}&=&\hspace{-2mm}\E_{\ppi(\state,\action|\btheta)}[\sum_{t=0}^{\infty}\gamma^{t}\rwtnew],\label{eq:J_new}
\end{eqnarray}}
for $\eta\geq0$ that controls the intrinsic reward. 
In this new reward, an agent's anticipation of the answer is taken into
account when updating the policy. When the seeker knows the answer
and its gain is small, the parameters are not changed significantly.
In other words, there is no need for further changes to the questions
where the answer is known. On the other hand, when the agent anticipates
a large gain from the answer and receives a large reward, the policy
has to be adjusted by a larger change in the parameters. Similarly,
if the agent expects a large gain and is not rewarded, there has to
be significant update in the policy.

In addition, for each parameter of the \seeker and each corresponding query $\actiont$, we have a
different gain. As such, when the variance of this gain $\mathbb{V}_{\actiont}[\gain(\statet,\actiont)]$ is small,
all the queries are expected to have similar answer and hence are almost the same. 

The advantage of this approach is twofold: (1)
it helps deal with sparse rewards and (2)
if a query's response  carries more information by providing better gain, we encourage its positive reinforcement.
%we encourage the method to ask the most informative questions.
 This allows the agent to learn to mimic the behavior of the
goal executor and generalize to unseen cases.

\subsection{\textbf{Information \Seeker\!\!'s Belief\label{subsec:question}}}
As discussed in Eq. (\ref{eq:sample_act}), each query is sampled from the \emph{seeker's policy distribution}. 
%of the parameters of the \seeker.
%Considering Eq.~\ref{eq:ans} and the need for the \ansr
%to consider the distribution of the policies,
%instead of finding a single policy as parameterized by $\btheta$,
%we model  the \emph{seeker's policy distribution}. 
Each
sample of the parameter $\btheta$ gives rise to a different querying
policy allowing us to model policy distribution.
This distribution
allows for the agent to consider alternatives, or contemplates, various
query policies to improve the overall dialog performance. As such,
here we consider the policy parameter $\btheta$ as a random variable
(leading to random policies that we can model their distribution)
and seek a distribution to optimize the expected return. We incorporate
a prior distribution $\ppip$ over the policy parameter, for instance, for
when we have no answer for query-response pairs or to incorporate
prior domain knowledge of parameters. 
The posterior in the conventional definition is
$$\ppi(\btheta|\histT,I) \propto \ppi(o|\{\actiont\}_t^T,\{\answert\}_t^T,\hist,I,\btheta)\ppip(\btheta).$$
where we denote $\histT=C\cup\{\{\actiont\}_t^T,\{\answert\}_t^T,o\}$ as an augmented context with the rollout of one seeker's round (for instance a dialog round in GuessWhat).
Since we need to define an additional likelihood for the goal and even then this posterior is intractable (unless major approximations and simplifying assumptions are made), we alternatively utilize the RL framework in which this posterior is used for. Specifically, we formulate the problem to find the policy distribution $\ppi$ under which the expected cumulative reward is maximized with additional prior regularization:
%This prior regularizes $\btheta$ for when no informative reward is present.
% We regularize the optimization of $\ppi$ as:
{\small\begin{equation}
\max_{\ppi}\bigg\{\E_{\ppi(\btheta|\histT,I)}[J(\btheta)]-\alpha\kl(\ppi\|\ppip)\bigg\},\label{Opt}
\end{equation}} %$\kl(\ppi\|\ppip)$,
%{\begin{align*}
%\text{where}\quad \kl(\ppi\|\ppip)=\E_{\ppi}[\log\ppi(\btheta|\histT,I)-\log\ppip(\btheta)].
%\end{align*}}
where $\kl(\ppi\|\ppip)=\E_{\ppi}[\log\ppi(\btheta|\histT,I)-\log\ppip(\btheta)]$.
Effectively we seek a parameter distribution that gives rise to policies
that maximize the expected reward while is close to the prior.
It is easy to see if we use an uninformative prior such as a uniform distribution, the second KL term is simplified to the entropy % $\mathbf{H}(\ppi)=\E_{\ppi(\btheta)}[-\log\ppi(\btheta)]$
of $\ppi$. Then optimization in Eq. (\ref{Opt}) becomes $\max_{\ppi}\Big\{\E_{\ppi(\btheta|\histT,I)}[J(\btheta)]+\alpha\mathbf{H}(\ppi)\Big\}$
which explicitly encourages exploration in the parameter space. % $\btheta$.
This exploration yields diverse policies that result in varied queries.

By taking the derivative of the objective function in Eq. (\ref{Opt})
and setting it to zero, the optimal distribution of policy parameter
$\btheta$ is obtained as
\begin{equation}
\ppi(\btheta|\histT,I)\propto\exp\left(J(\btheta)/{\alpha}\right)\ppip(\btheta).\label{eq:Bayes}
\end{equation}

In this formulation, $\ppi(\btheta|\histT,I)$ is the ``posterior'' of the parameters $\btheta$ in the conventional
Bayesian approach. Then, $\exp(J(\btheta)/\alpha)$ is effectively the ``likelihood\textquotedbl{}
function. The coefficient
$\alpha$ is the parameter that controls the strength of exploration
in the parameter space and how far the posterior is from the prior.
As $\alpha\rightarrow0$, samples drawn from $\ppi(\btheta|\histT,I)$
will be concentrated on a single policy %around the optimum of $\E[J(\btheta)]$
and lead to less diverse seekers.

Remember from Eq. (\ref{eq:J_new}) that the ``likelihood'' here
% in the estimation of the posterior
considers the agent's anticipation
of the answer. If its reward is higher, then a larger change to the
parameter is needed to allow exploitation of new knowledge about the
effect of the current policy on the goal.

Similar ideas of entropy regularization has been investigated in other
reinforcement learning methods \cite{svpg,trpo}. However,
in our approach we use the regularization to obtain the posterior
for the policy parameters in the information seeking framework where
the gain from the \ansr refines the policy distribution.

\subsection{\textbf{\Seeker\!\!'s Posterior Update\label{subsec:refinement}}}
A conventional method to utilise the posterior in Eq. (\ref{eq:Bayes})
is Markov Chain Monte Carlo (MCMC) sampling. However, MCMC
methods are computationally expensive, suffer from slow convergence
and have high-variance due to stochastic nature of estimating $J(\btheta)$.
Since estimating $J(\btheta)$ by itself is a computationally demanding
task and may vary significantly for each policy, we look for an efficient
alternative. Thus, rather than $J(\btheta)$, we use the gradient
information $\nabla_{\btheta}J(\btheta)$ that points to the direction
for seeker's policy change using the \emph{Stein variational gradient
descent} (SVGD) for Bayesian inference \cite{svgd_gf,svgd}. SVGD
is a nonparametric variational inference algorithm that leverages
efficient deterministic dynamics to transport a set of particles $\{\btheta_{i}\}_{i=1}^{n}$
to approximate given target posterior distributions $\ppi(\btheta|\histT,I)$.
Unlike traditional variational inference methods, SVGD does not confine
 the approximation within a parametric families, which means the \seeker\!\!'s
policy does not need to be approximated by another. Further, SVGD
converges faster than MCMC due to the deterministic updates that efficiently
leverage gradient information of the \seeker\!\!'s policy posterior. This
inference is  efficiently performed by iteratively updating multiple
``particles'' $\{\btheta_{i}\}$ as $\btheta_{i}=\btheta_{i}+\epsilon_\theta\psi^{*}(\btheta_{i})$,
where $\epsilon_\theta$ is a step size and $\psi(\cdot)$ is a function
in the unit ball of a reproducing kernel Hilbert space (RKHS). Here,
$\psi^{*}$ is chosen as the solution to minimizing KL divergence
between the particles and the target distribution. It was shown
 that this function has a closed form empirical estimate \cite{svgd}:
 \begin{figure}[t]
	\centering
	\includegraphics[width=0.6\columnwidth]{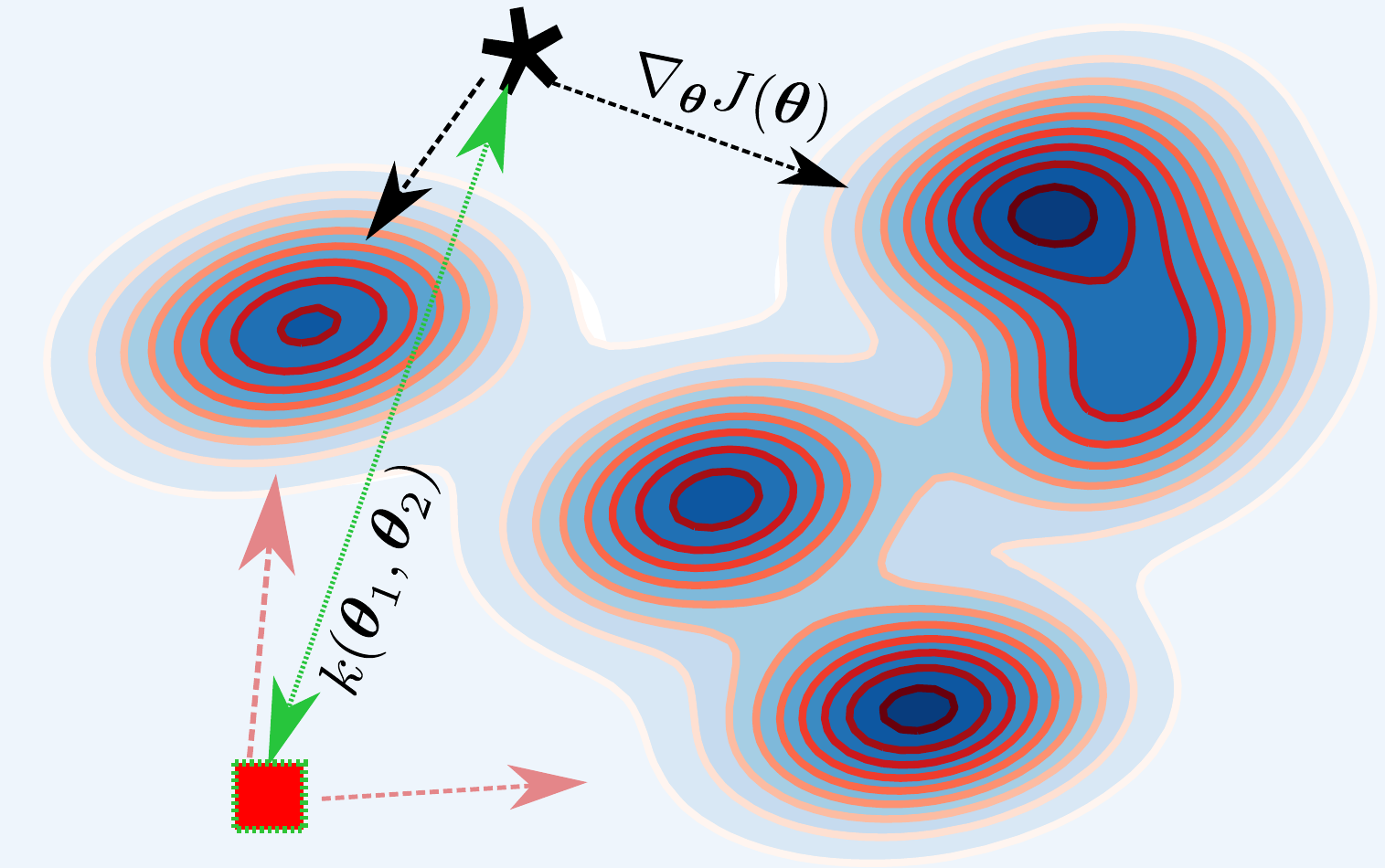}
	\caption{\small{An illustration of the multi-modal distribution of the \seeker\!\!s policies. Unlike conventional policy gradient methods that only explore nearest mode, our novel approach always use a number of initial points (i.e. the policy parameters) to explore multiple modes collaboratively in analogy of human contemplation of multiple strategies. We only show two initial points, a red rectangle and black asterisk, for the ease of visualization.}}\vspace{-3mm}\label{fig:algo_plot}
\end{figure}
% \begin{equation}
% \psi^{*}=\max_{\psi\in\H}\big\{-\frac{d}{d\epsilon}\kl(\rho_{[\epsilon\psi]}\|\ppi),\qquad\text{s.t.}\quad||\psi||_{\H}\leq1\},
% \end{equation}
% where $\rho_{[\epsilon\phi]}$ denotes the distribution of $\btheta'=\btheta+\epsilon\psi(\btheta)$,
% and the distribution of $\btheta$ is $\rho$. This optimization has
% a closed-form solution \cite{svgd}:
% \[
% \psi^{*}(\btheta)=\E_{\bvartheta\sim\rho}[\nabla\log\ppi(\bvartheta)]k(\bvartheta,\btheta)+\nabla_{\bvartheta}k(\bvartheta,\btheta)],
% \]
% where $k(\bvartheta,\btheta)$ is the positive definite kernel associated
% with the RKHS space.
% The empirical estimate for current particles
% $\{\btheta_{i}\}$ using the is the Stein variational gradient, i.e.
\vspace{0mm}{\small\begin{align}
 \hat{\psi}(\btheta_{i})&=\frac{1}{n}\sum_{j=1}^{n}[\nabla_{\btheta_{j}}\log\ppi(\btheta_{j}|\histT,I)k(\btheta_{j},\btheta_{i})\nonumber \\
 &\qquad\qquad\qquad\qquad\qquad\quad+\nabla_{\btheta_{j}}k(\btheta_{j},\btheta_{i})].
\label{eq:svgd}
\end{align}}
where $k$ is the  the positive definite kernel associated with the RKHS space.
In this update rule $\hat{\psi}$, the first term involves the gradient
$\nabla_{\btheta}\log\ppi(\btheta|\histT,I)$ which moves the seeker's policy
particles $\btheta_{i}$ towards the high probability regions by sharing
information across similar particles. The second term $\nabla_{\btheta_{j}}k(\btheta_{j},\btheta_{i})$
utilizes the curvature of the parameter space to push the particles
away from each other, which leads to diversification of the seeker's
policies.

An example of the landscape
of the policies is shown in Fig.~\ref{fig:algo_plot}. Each
initial sample from the policy distribution can move towards one of the modes
of a highly multi-modal distribution.
These moves are
governed by the gradient of the policy that in our case consists of the
agent's belief about the answer and its consequence once its response
is known. In addition, kernel $k$ controls the distance between
the parameters to deter from collapsing to a single point in
multi-modal distribution.
It is intuitive from the figure that a better gradient from
the rewards by incorporating the answers and considering the
distribution of policies improves the performance of the seeker by
guiding the parameter updates.

It is noteworthy to mention that even though we only receive the reward for one 
sample of $\actiont$ taken from one particle (from Eq. \ref{eq:ans_gain}), due to our formalization the 
posterior is adjusted for all particles allowing the feedback to be propagated.

It is important to note that diversification in the parameter space
allows for an accurate modeling of a highly multimodal policy space.
Otherwise, the policy distribution collapses to a single point which
is the same as the conventional maximum a posteriori (MAP) estimate.
This MAP estimate only considers a single policy that in the highly complex
task of vision-language is inadequate.

{\begin{algorithm}[t]
\caption{Seeker}
\label{alg:algexample}
\begin{algorithmic}
\State \textbf{Input}:
Learning rate $\epsilon_\theta,\epsilon_\omega$, kernel $k(\btheta,\btheta')$, %prior $q_0(\btheta)$,
initial policy particles $\{\btheta_{i}\}$, context history $\hist$, image $I$.
\For{iteration $t=0,1,..,T$}
\For{particle $i=1,\ldots,n$}
	\State Sample $\action\sim \ppi(\action|\statet,\hist, I;\btheta_{i})$
	\State Compute $\gain(\statet,\action)$ from Eq.~\eqref{eq:ans_gain}. %~\eqref{eq:ucb_bound}.
\EndFor
	\State{Select $\actiont$ with maximum gain}
	\State $\answert= \arg\max_a p(\answer|\statet,\actiont,\hist, I;\bomega)$ 
	\State $\bomega\gets\bomega+\epsilon_\omega\nabla_\bomega\log\Big(p(\y|\answert,\statet;\bomega)\Big)$\Comment{score}
	% \State Compute new rewards $r^{\text{new}}$ from Eq.~\eqref{eq:newreward}.
	\State Compute $\nabla_{\btheta_{i}}J(\btheta_{i})$ in Eq.~\eqref{eq:J_new}. \Comment{from Eq.~\eqref{eq:newreward}}
	% \State $\Delta\bomega\gets\Delta\bomega+$
\For{particle $i=0,1,..,n$}
	\State $J_\text{new}(\btheta_j)=\frac{1}{\alpha}J(\btheta_{j})+\log\ppip(\btheta_{j})$
	\State $\footnotesize{\Delta\btheta_{i}\!\gets\!\frac{1}{n}\sum_{j=1}^{n}\Big[\nabla_{\btheta_{j}}J_\text{new}(\btheta_j)k(\btheta_{j},\btheta_{i})+\nabla_{\btheta_{j}}k(\btheta_{j},\btheta_{i})\Big]}$
	\State $\btheta_{i}\gets\btheta_{i}+\epsilon\Delta\btheta_{i}$ \Comment{update the policy}
\EndFor
 \EndFor \end{algorithmic}
\end{algorithm}}

\begin{comment}
Note that the answer is predicted by our model is
\begin{align}
\pans(\answert|\statet,\hist,I;\bomega)&=\int p(\answert|\actiont,\hist,I,\statet;\bomega)\label{eq:ans} \\
&\qquad\times\ppi(\actiont|\statet,\hist,I;\btheta)\ppi(\btheta|\hist,I)d\btheta. \nonumber
\end{align}
Here, we have taken the distribution of $\btheta$ to consider all the policies for queries in the \seeker.
\end{comment}

%\begin{wrapfigure}{r}{0.45\linewidth}
	%\centering
	%\hspace{-11mm}\includegraphics[width=1.15\linewidth]{img/algorithm_plot.pdf}
	%\hspace{-15mm}\vspace{-33mm}
	%\caption{\small An illustration of the multi-modal distribution of the \seeker\!\!s policies. Unlike conventional policy gradient methods that only explore nearest mode, our novel approach always use a number of initial points (i.e. the policy parameters) to explore multiple modes collaboratively in analogy of human contemplation of multiple strategies. We only show two initial points, a red rectangle and black asterisk, for the ease of visualization.}\label{fig:algo_plot}\vspace{-10mm}
%\end{wrapfigure}

\section{Experiments}
To evaluate the performance of the \seeker we conducted experiments
on two different goal-oriented vision-and-language datasets: GuessWhat~\cite{guesswhat_game} and CLEVR~\cite{clevr}.
The former is a visual dialog task while the later is a compositional visual question answering task.
%The former is a c'operative game in which the
%dialogue between two agents is carried out to achieve a common goal
%while the later is a semi-cooperative game where two agents have adversarial
%objectives.
In both experiments we pre-train the networks using the supervised
model and refine using reinforcement learning, as is common practice in this area \cite{visdial_rl,guesswhat_game}.
%A Without using supervised learning first, the dialog model may diverge from human language. 
Policies are generated by sampling from the policy posterior $
\btheta\sim\ppi(\btheta|\histT,I)$ and generate the query 
with the highest gain measured by the \ansr.
Our approach outperforms the previous state-of-art in both 
cases.
Note that our approach is architecture neutral and as such we expect using better representations to even improve performance further. %Moreover, the reward for the  information seeking agent is $1$ when the goal is achieved, otherwise $0$. 
%ur approach outperforms the baseline and previous state-of-art in both cases. In
%both experiments we pre-train the networks using the supervised
%model
%A which is then used in the reinforcement learning for further refinement.
%and refine using reinforcement learning.
%/A
%To that end, we employ a two stage algorithm in which
%we learn to imitate the human dialogue behaviour in a supervised learning
%task and subsequently fine-tune for better generalisation and goal
%discovery using reinforcement learning. Similar two-stage approaches
%are taken in \cite{visdial_rl,guesswhat_game,dealornodeal}. Without
%using supervised learning first, the dialogue model may diverge from
%human language.

%%%%%%%%%%%%%%%%%%%%%%%%%%%%%%%%%%%%%%%%%%%%%%%%%%%%%%%%%%%%%%%%%%%%%%%%%%%%
%%%%%%%%%%%%%%%%%%%%%%%%%%%%%%%%%%%%%%%%%%%%%%%%%%%%%%%%%%%%%%%%%%%%%%%%%%%%

\subsection{\vspace{0mm}GuessWhat}

GuessWhat \cite{guesswhat_game} is a classical goal-oriented visual dialog game.
In each game, a random object in the scene is assigned to the answerer, but hidden from the questioner (our \Seeker). The questioner can ask a series of yes/no questions to
locate the object. The list of objects is also hidden from the
questioner during the question-answer rounds. Once
the questioner has gathered enough information, the
guesser (our \exec) can start to guess. 
If a guess is correct the game is successfully concluded.
The extrinsic reward is "one" at the end of a dialog (i.e. series of question-answers) when the predicted object matches the true object the oracle chose. 
%The dataset includes $155,281$ dialog of
%$821,955$ pairs of question/answers with vocabulary size $11,465$ on
%$66,537$ unique images and $134,074$ objects.

%The vocabulary size is $1,192$.
 %\footnote{\url{https://guesswhat.ai}}.

\vspace{-10pt}
\paragraph{Implementation Details}
%We follow the same experimental setup as \cite{guesswhat_game} in
%which three main components are built: a yes/no \texttt{executor} agent, a guesser \texttt{answerer} agent
%and a questioner or information \texttt{seeker}.
In our model, the information seeker is a set of $10$ recurrent neural
networks (RNNs) that represent the particles from the likelihood in Eq.~(\ref{eq:Bayes}). We use LSTM \cite {lstms} cells
in these RNNs for which the parameters are updated according to Eq.~(\ref{eq:svgd}) to simulate the posterior.
The hidden representations of these LSTM networks (with size $1024$) correspond to the state in the reward function. 
% To obtain a distribution over tokens, a softmax is applied to this output.
The image representation is obtained using VGG~\cite{vgg}.
The concatenation of the image and history features are given to each particle in the \seeker for question generation where each word is sampled conditioned on its
previous word.
We use $u(\cdot)=\exp(\cdot)$ to operate as the score function for computing gain in Eq.~(\ref{eq:ans_gain}), and  reward in Eq.~(\ref{eq:J_new}).

We set $\eta=0.1\times\frac{\text{epoch}_{\max}-\text{epoch}}{\text{epoch}_{\max}}$ to encourage the policies to explore more in the initial stages. In addition, $\alpha=0.001$. We use the median trick from \cite{svgd} to compute the RBF-kernel's hyper-parameter which essentially ensures $\sum_{j}k(\btheta_i,\btheta_j)\approx 1$.

%\textcolor[rgb]{1.00,0.00,0.00}{We set $\eta$, $\beta$ and the RBF-kernel's hyper-parameter similar to
%the experiments in CLEVR, however we set $\alpha=0.001$ here using grid-search.}

% Once the seeker is trained using RL, we take three approaches
% to evaluating the performance of the seeker: (1) \textit{sampling} where
% the subsequent word is sampled from the multinomial distribution in
% the vocabulary, (2) \textit{greedy} where the word with maximum probability
% is selected and (3) \textit{beam search} keeping the K-most promising
% candidate sequences at each time step (we choose $K=20$ in all
% experiments). During training the baseline uses the greedy approach
% to select the sequence of words as in \cite{guesswhat_game}.
% %Beam search is intended to tackle the inherent uncertainty in the
% %dialogue by exploring the consequence of generating a sequence of words.

\vspace{-10pt}
\paragraph{Overall Results}
%A We can compare the results from the objects in the training set (\textbf{New Object}) where the object is new in an already seen image or the test set where the image is unseen (\textbf{New Image}).
We compare two cases, labeled \textit{New Object} and \textit{New Image}.  In the former the object sought is new, but the image has been seen previously.  In the latter the image is also previously unseen.
We report the prediction
accuracy for the guessed objects. It is clear that the accuracies
are generally higher for the new objects as they are obtained from
the already seen images.

The results are summarized in Table \ref{tbl:guesswhat1}. As shown,
using the conventional REINFORCE \cite{strub2017end} by either sampling
each word (RL-S) or greedily selecting one (RL-G) improves the performance compared to the supervised baseline significantly. Since our approach
explore and exploits the space of policies for question generation better, it
achieves better performance. Furthermore, this performance is improved when
a better goal seeker or \ansr model is employed. Better \ansr leads
to more realistic intrinsic rewards that corresponds to true gains and guide the
policy distribution to a better posterior. For instance, employing a Memory network
\cite{memorynet} within the \ansr improves its performance that in turn is reflected in
the quality of the questions and consequently agent's ability to achieve goals more accurately.
Note that even in this case the single particle experiment has improved since the rewards are more 
accurate to evaluate the question-answer relations.
% The major improvement is due to better policy update as a consequence of better reward (including the intrinsic one).

% Note that our approach is generic enough that can be used in combination with other
% neural architectures that represent the data better.

\begin{table}
\centering{}{\footnotesize{}}%
\resizebox{0.99\linewidth}{!}{
\small
\begin{tabular}{l|c|c}
\hline
 {\footnotesize{}\textbf{Model}}&\multicolumn{1}{c|}{{\footnotesize{}\textbf{New Object}}} &
\multicolumn{1}{c}{{\footnotesize{}\textbf{New Image}}}\tabularnewline
\hline
 {\footnotesize{}Supervised-S}~\cite{guesswhat_game}
 & {\footnotesize{}$41.6$} & {\footnotesize{}$39.2$} \tabularnewline
 {\footnotesize{}Supervised-G}~\cite{guesswhat_game}
 & {\footnotesize{}$43.5$} & {\footnotesize{}$40.8$} \tabularnewline
{\footnotesize{}RL-S}~\cite{strub2017end} & {\footnotesize{}$56.5$} & {\footnotesize{}$58.5$} \tabularnewline
{\footnotesize{}RL-G}~\cite{strub2017end} & {\footnotesize{}$60.3$} & {\footnotesize{}$58.4$} \tabularnewline
{\footnotesize{}Tempered}~\cite{tempered} & {\footnotesize{}$62.6$} & -\tabularnewline
{\footnotesize{}Tempered-Seq2Seq}~\cite{tempered} & {\footnotesize{}$63.5$} & -\tabularnewline
{\footnotesize{}Tempered-MemoryNet}~\cite{tempered} & {\footnotesize{}$68.3$} &- \tabularnewline
\hline
{\footnotesize{}Ours (no intrinsic reward)} & {\footnotesize{}${63.3}$} & - \tabularnewline
{\footnotesize{}Ours} & {\footnotesize{}${64.2}$} & {\footnotesize{}${62.1}$}\tabularnewline
{\footnotesize{}Ours+MemoryNet (Single)} & {\footnotesize{}${70.1}$} &
{\footnotesize{}${67.9}$}\tabularnewline
{\footnotesize{}Ours+MemoryNet} & {\footnotesize{}$\boldsymbol{74.4}$} &
{\footnotesize{}$\boldsymbol{72.1}$}\tabularnewline
\hline
\end{tabular}}
\vspace{-4pt}
\caption{\small{Accuracy in identifying the goal object in the GuessWhat dataset (higher is better).
The "S" indicator is for sampling for words method vs "G" which is greedy.  Ours+MemoryNet is the method with modified \ansr that employs Memory network and Attention. Further, (Single) indicates training our method with a single particle.}}
\label{tbl:guesswhat1}
\vspace{-7pt}
\end{table}

%%%%%%%%%%%%%%%%%%%%%%%%%%%%%%%%%%%%%%%%%%%%%%%%%%%%%%%%%%%%%%%%
%%%%%%%%%%%%%%%%%%%%%%%%%%%%%%%%%%%%%%%%%%%%%%%%%%%%%%%%%%%%%%%%

\subsection{\vspace{0mm}CLEVR}

\begin{figure*}
\centering
\vspace{-4mm}
\includegraphics[width=0.84\linewidth]{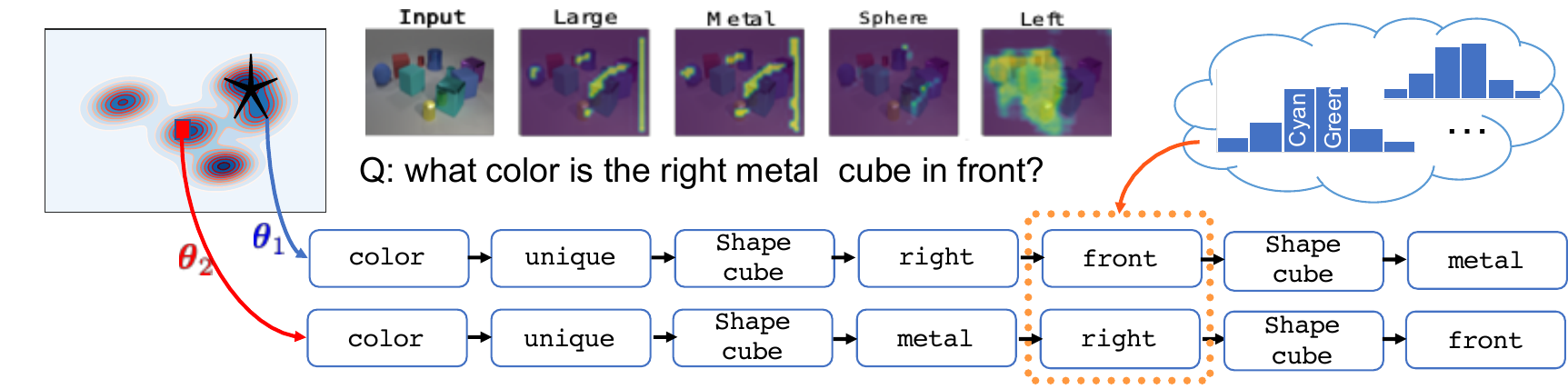}
\caption{\small{An example of a question and the programs generated using samples from the posterior in CLEVR.
Samples from the policy distribution take the input image and the question and generate
its corresponding programs. As observed, these two samples produce different program sequences which enable to explore multiple distributions over the goal (final answer) shown on top in the cloud. Expected score of each question $\gain_{\bomega}(\statet,\actiont)$ gives us an indication of which one is better to ask.
% The fist row indicates the input image and the corresponding attention when generating
% program for the first question. The programs are generated by samples from the posterior.
% For each question, the red boxes are the functions for each program and the blue ones  are the answers.
% It is interesting to note that for the second
% question, different policy samples of our approach generate different programs, albeit the expected
% generated program queries shape and count.
}}\label{fig:clevr_all}
\vspace{-2mm}
\end{figure*}

CLEVR \cite{clevr} is a synthetically generated dataset containing 700K (image,  question,  answer,  program)  tuples.
Images are  3D-rendered  objects of various shapes,  materials,  colors,  and sizes.
Questions are compositional in nature and range from counting questions to comparison questions and can be 40+ words long. An answers is a word from a set of $28$ choices.
For each image and question, a program consists of step-by-step instructions,
on how to answer the question. During the test, the programs are not given, which need to be generated conditioned on the input question.
The extrinsic reward is "one" when the generated program yields the correct answer.

{
\begin{table}[t]
  \centering
  \resizebox{1.03\linewidth}{!}
  {
  \hspace{-3mm}\begin{tabular}{lcccccc}
    \toprule
    \multirow{2}{*}{Model} &
    \multirow{2}{*}{Overall} & \multirow{2}{*}{Count}\hspace{-5mm}& \hspace{-2mm}Compare&
    \multirow{2}{*}{Exist} \hspace{-5mm}& Query \hspace{-5mm}& \hspace{-2mm}Compare\hspace{-3mm}\\
    \hspace{-3mm} & \hspace{-3mm}& \hspace{-3mm}& \hspace{-3mm}Numbers& & Attribute & \hspace{-2mm}Attribute\hspace{-5mm}\\
    \midrule
    NMN \cite{nmn}\hspace{-3mm}             & 72.1 & 52.5 & 72.7 & 79.3 & 79.0 & 78.0\\
    N2NMN \cite{hu2017learning}\hspace{-3mm}      & 88.8 & 68.5 & 84.9 & 85.7 & 90.0 & 88.8\\
    \midrule
    Human \cite{clevr}\hspace{-3mm}           & 92.6 & 86.7 & 86.4 & 96.6 & 95.0 & 96.0\\
    \midrule
    LSTM+RN \cite{relational-reasoning}\hspace{-3mm} & 95.5 & 90.1 & 93.6 & 97.8 & 97.1 & 97.9\\
    PG+EE (9k) \cite{infering}\hspace{-6mm}     & 88.6 & 79.7 & 79.7 & 89.7 & 92.6 & 96.0\\
    PG+EE (18k) \cite{infering}\hspace{-6mm}    & 95.4 & 90.1 & 96.2 & 95.3 & 97.3 & 97.9\\
    PG+EE (700k) \cite{infering}\hspace{-5mm}     & 96.9 & 92.7 & 98.6 & 97.1 & 98.1 & 98.9\\
    FiLM \cite{film}\hspace{-3mm}           & 97.6 & 94.5 & 93.8 & 99.2 & 99.2 & 99.0\\
    DDRprog \cite{ddrprog}\hspace{-3mm}       & 98.3 & 96.5 & 98.4 & 98.8 & 99.1 & 99.0\\
    MAC \cite{cans}\hspace{-3mm}          & 98.9 & 97.2 & 99.4 & 99.5 & 99.3 & 99.5\\
    TbD-net \cite{transparancydesign}\hspace{-3mm}  & 98.7 & 96.8 & 99.1 & 98.9 & 99.4 & 99.2\\
    TbD-net++ \cite{transparancydesign}\hspace{-3mm} & 99.1 & 97.6 & 99.4 & 99.2 & 99.5 & 99.6\\
    \midrule\midrule
    Ours+G+entropy (9k) \hspace{-3mm}      & 91.4 & 86.4 & 93.6 & 89.8 & 93.2 & 96.2\\
    Ours+G+entropy (18k) \hspace{-3mm}     & 95.6 & 93.3 & 96.8 & 95.4 & 97.8 & 98.1\\
    Ours+G+entropy (700k)\hspace{-3mm}     & 97.4 & 96.8 & 98.1 & 98.2 & 96.2 & 98.1\\
    Ours+D+entropy (9k) \hspace{-3mm}     & 94.7 & 92.2 & 95.6 & 93.2 & 95.1 & 97.7\\
    Ours+D+entropy (18k) \hspace{-3mm}    & 96.6 & 94.6 & 96.1 & 95.6 & 98.1 & 98.6\\
    Ours+D+entropy (700k)\hspace{-3mm}    & 98.3 & 98.1 & 99.1 & 97.1 & 98.6 & 98.8\\
    \midrule
    Ours+G+exp (9k) \hspace{-3mm}      & 91.8 & 87.5 & 93.7 & 90.2 & 93.1 & 96.5\\
    Ours+G+exp (18k) \hspace{-3mm}     & 96.3 & 93.3 & 96.8 & 95.4 & 97.8 & 98.1\\
    Ours+G+exp (700k)\hspace{-3mm}     & 98.0 & 96.2 & 98.6 & 98.0 & 98.0 & 99.0\\
    Ours+D+exp (9k) \hspace{-3mm}     & 95.2 & 91.5 & 96.7 & 93.8 & 95.7 & 98.7\\
    Ours+D+exp (18k) \hspace{-3mm}    & 97.1 & 94.5 & 98.2 & 96.1 & 98.3 & 98.6\\
    Ours+D+exp (700k)\hspace{-3mm}    & 98.9 & 97.8 & 99.2 & 98.9 & 99.5 & 99.3\\
    Ours+D+exp++ (700k)\hspace{-5mm}  & \textbf{99.2} & \textbf{97.8} & \textbf{99.5} & \textbf{99.4} & \textbf{99.6} & \textbf{99.6} \\
    \bottomrule
  \end{tabular}}
  \caption[CLEVR Dataset Accuracy]{{Performance comparison of state-of-the-art models on the CLEVR
    dataset. "Ours+G+entropy" is our seeker when used with the generic architecture and entropic gain; "Ours+D+entropy" is the same except for using designed architecture. Similarly, "Ours+G+exp" is generic architecture with $u_{\text{exp}}$; and, "Ours+D+exp" is its designed counterpart. We achieve state of the
    art performance, especially using smaller ground-truth programs. The `++' indicator shows a model was trained using higher-resolution $28\times28$ feature maps rather than $14\times14$.}}\label{tbl:clevr}\vspace{-3mm}
\end{table}}

\vspace{-10pt}
\paragraph{Implementation Details}
We follow the experimental setup of \cite{infering,clevr} in which uses a ResNet \cite{resnet} to encode
the given images and a standard LSTM \cite{lstms}  to generate programs in the context encoder.
 %For the \exec and \ansr we use a modular network \cite{nmn}.
%For simplicity we use a two-stage scenario where  we first pre-train all the networks end-to-end
%using the supervision signal from the training triplets $(I, \actiont, \ansopt)$ and then
%use RL to fine-tune.
We use $10$ particles in Algorithm~\ref{alg:algexample} to model the policy distribution using
samples from the pre-trained model with added noise so that they correspond
to different initial policies. For more efficient implementation, we use two
 sets of shared parameters for the encoder in the underlying Seq2Seq
model and use independent parameters for the LSTM decoder. This parameter sharing also ensures
there are common latent representations that particles learn.
We use
our information \seeker model in Section~\ref{sec:infoseeker_oracle} to generate samples or
programs for each question and consider the consequence of that program using the \ansr internally to choose one.
Once a program is generated, it is then executed by the goal \exec to obtain the feedback and
compute the corresponding rewards. The computed reward is then used to update the policy distributions as discussed.
We use the Adam optimization
method with learning rate set to $10^{-5}$ to update both the \seeker and the \ansr's parameters.
The testing procedure thus takes an
image and question pair, produces a program, then the goal executor produces an
answer. The goal \exec then evaluates the quality of the generated program. We set $\alpha=0.01$
and $\eta$ similar to the GuessWhat experiment.

% {
% \begin{table}
%   \centering
%   \resizebox{1.03\linewidth}{!}
%   {
%   \hspace{-3mm}
%   \begin{tabular}{llr}
%   \multicolumn{3}{c}{\includegraphics[width=\linewidth]{img/clevr}}\\
%   \toprule
%   {\footnotesize"Is there a large metal sphere on left?"} &
%   &
%   {\footnotesize\color{blue}{No}}\\
%   {\footnotesize"How many cubes?"} &
%       {\footnotesize\textt{count $\to$ filter(cube)}} &
%       {\footnotesize\color{blue}{5}}\\
%     \bottomrule
%   \end{tabular}}
%   \caption[a]{a}\label{tbl:clevr_img}
% \end{table}}

\vspace{-10pt}
\paragraph{Overall Results}
 For the \ansr and the \exec, we consider two alternative baselines:
 (G)eneric similar to \cite{infering} where each module follows a generic architecture; and,
 (D)esigned similar to \cite{transparancydesign} where each module is specifically designed based on the
desired operation.
%\begin{wrapfigure}{r}{0.45\linewidth}
	%\vspace{-1mm}\hspace{-8mm}
	%\includegraphics[width=1.19\linewidth]{img/param_dist_rewards.pdf}
	%\caption{Average reward for the agent at each iteration (Top); and,
		%average distance between particles in the posterior for CLEVR (Bottom).}\label{fig:clevr_param_dists}
	%\vspace{-2mm}
%\end{wrapfigure}
\begin{figure}[htb]
	\centering
	\vspace{-3mm}\hspace{-2mm}
		\includegraphics[width=0.51\columnwidth]{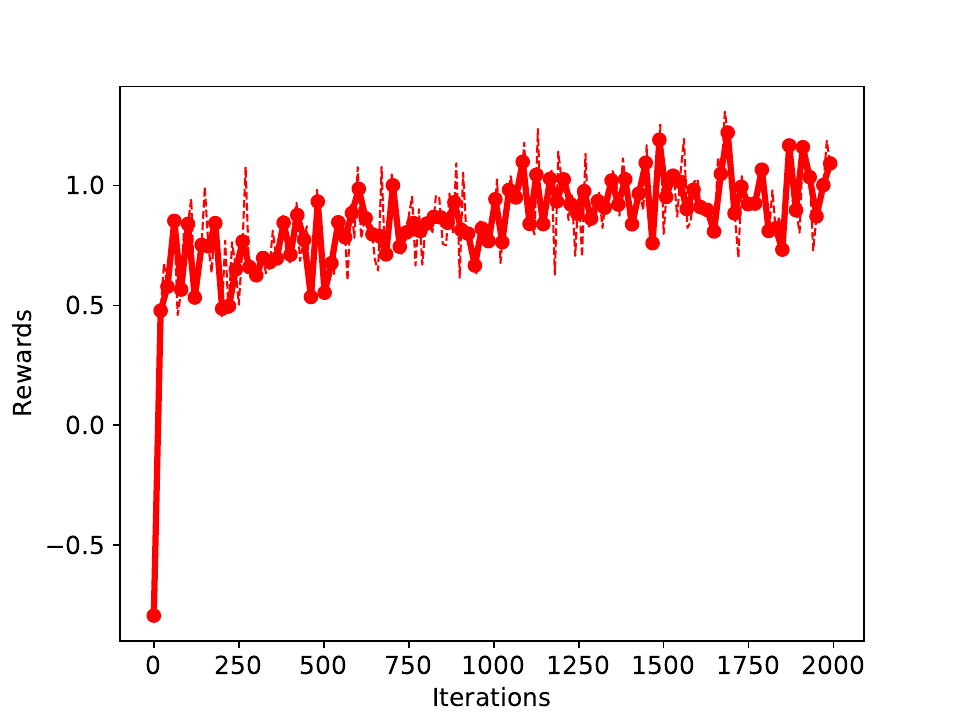}\hspace{-3mm}
		\includegraphics[width=0.51\columnwidth]{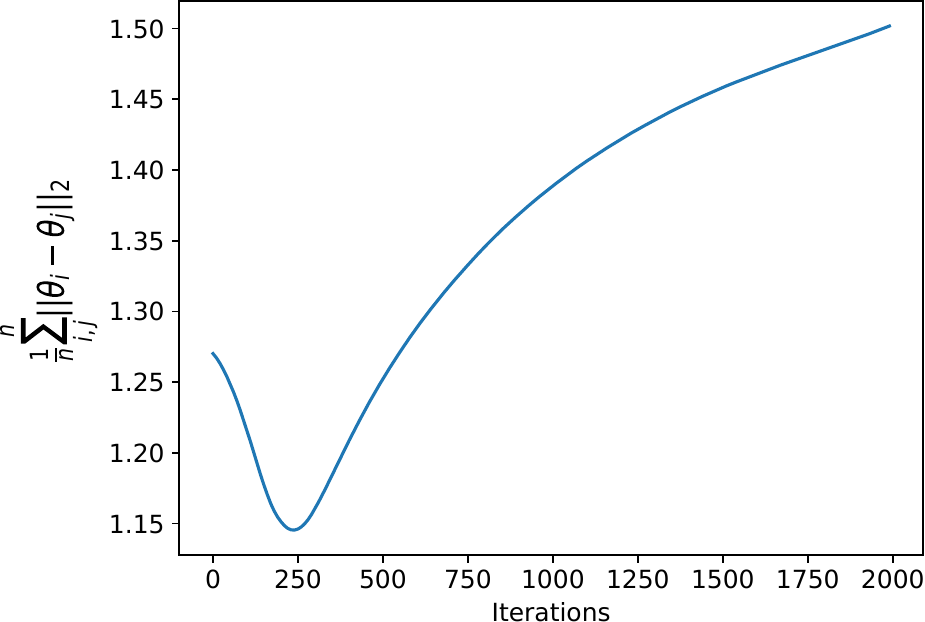}
	\caption{\small{Average reward for the agent at each iteration; and,
		average distance between particles in the posterior for CLEVR.}}\vspace{-1mm}\label{fig:clevr_param_dists}
\end{figure}

We report the accuracy of the goal executor. Since the later case provides a better representation on each module, we expect it to perform better. %In addition since the module networks are utilised with attention mechanisms, we are able to visualise the attended areas.
Further, we use two functions $u_\text{entropy}(\cdot)=\log(\cdot)$ (corresponding to the information-theoretic notion of gain in the expectation) and $u_\text{exp}(\cdot)=\exp(\cdot)$ to operate on the output of the \ansr's score to compute the gain and ultimately the new reward in Eq.~(\ref{eq:ans_gain}) and~(\ref{eq:J_new}).
The results in Tab.~\ref{tbl:clevr} 
%A As shown our approach outperforms the baselines by a significant margin.
show that our approach outperforms the baselines almost to the maximum extent possible.
%/A
In particular, our approach almost achieves the same performance as that of \cite{infering} with half the programs used for training with the same neural architecture. Moreover, the choice of $u$ affects the policies found, for instance using  $u_\text{entropy}$ generally leads to outperforming in the "count" function.
 Thus, since  $u_{\exp}$ has a smaller range and is smoother, it provides a more uniform penalty for mistakes in all modules during training leading to generally better performance.
%The samples from the posterior is used to generate the programs $\btheta\sim\ppi(\btheta|\ansopt,\actiont,\hist,I)$ by averaging over the scores given to each function.
As shown in Figure \ref{fig:clevr_all}, each sample from the policy generates a different program. In addition, we are able to utilize the attention mechanism in the model to \emph{reason} about where in the image the information seeker focuses.

Fig. \ref{fig:clevr_param_dists} plots the average reward at each iteration, and the average distance between the
particles in the policies.
If the problem was indeed unimodal (as conventional methods assume),
all the particles would collapse to a single point indicated by a zero average distance.
 However,
as is observed, while the distance between the particles decreases in early stages, they
soon increase indicating convergence to independent modes.
Our context encoder, unlike \cite{answer_questioner_mind}, is a pixel level model that does not extract objects explicitly from the given image. To be fair, we only consider methods that are directly comparable.

\section{Conclusion}
The ability to identify the information needed to support a conclusion, and the actions required to obtain it, is a critical capability if agents are to move beyond carrying out low-level prescribed tasks towards achieving flexible high semantic level goals.  The method we describe is capable of reasoning about the information it holds, and the information it will need to achieve its goal, in order to identify the action that will best enable it to fill the gap between the two.  Our approach thus actively seeks the information it needs to achieve its goal on the basis of a model of the uncertainty in its own understanding.

%If we are to enable agents that can interact flexibly with humans to achieve board, and loosely specified goals, the capability our approach demonstrates will be critical.
If we are to enable agents that actively work towards a high-level goal the capability our approach demonstrates will be critical. In particular, agents need to be able to consider alternative policies for achieving a goal and their corresponding uncertainty, evaluate the outcome of executing those policies and the information it gains. 

%\par\textbf{Acknowledgment:}
%This material is based on research sponsored by Air Force Research Laboratory and DARPA under agreement number FA8750-19-2-0501. The U.S. Government is authorized to reproduce and distribute reprints for Governmental purposes notwithstanding any copyright notation thereon.

\newpage
\bibliographystyle{plain}
\bibliography{lib}

\end{document}